\documentclass[10pt,twocolumn,letterpaper]{article}

\usepackage{cvpr}
\usepackage{times}
\usepackage{epsfig}
\usepackage{graphicx}
\usepackage{amsmath}
\usepackage{amssymb}

\usepackage{gensymb}
\usepackage{color, colortbl}
\definecolor{LightCyan}{rgb}{0.88,1,1}
\usepackage{pifont}
\newcommand{\cmark}{\ding{51}}%
\newcommand{\xmark}{\ding{55}}%

\usepackage[pagebackref=true,breaklinks=true,letterpaper=true,colorlinks,bookmarks=false]{hyperref}

\cvprfinalcopy 

\def\cvprPaperID{****} 

\ifcvprfinal\fi
\begin{document}

\title{3D-FCT: Simultaneous 3D object detection and tracking using feature correlation}

\author{Naman Sharma\\
Seagate Research\\
{\tt\small naman.sharma@seagate.com}
\and
Hocksoon Lim\\
Seagate Research\\
{\tt\small hocksoon.lim@seagate.com}
}

\maketitle

\begin{abstract}
   3D object detection using LiDAR data remains a key task for applications like autonomous driving and robotics. Unlike in the case of 2D images, LiDAR data is almost always collected over a period of time. However, most work in this area has focused on performing detection independent of the temporal domain. In this paper we present 3D-FCT, a Siamese network architecture that utilizes temporal information to simultaneously perform the related tasks of 3D object detection and tracking. The network is trained to predict the movement of an object based on the correlation features of extracted keypoints across time. Calculating correlation across keypoints only allows for real-time object detection. We further extend the multi-task objective to include a tracking regression loss. Finally, we produce high accuracy detections by linking short-term object tracklets into long term tracks based on the predicted tracks. Our proposed method is evaluated on the KITTI tracking dataset where it is shown to provide an improvement of 5.57\% mAP over a state-of-the-art approach.
\end{abstract}

\section{Introduction}\label{sec:intro}
3D object detection is a fundamental task required for a multitude of robot applications such as autonomous driving, object manipulation and augmented reality. LiDAR sensors are commonly utilized in autonomous driving scenarios which capture 3D information about the environment, and the aim of the 3D object detection task is to utilize this information to provide semantically labeled 3D {\em oriented} bounding boxes for all objects in the environment. A related task is 3D object tracking, which requires ID assignment for each detected object across time.

Most 3D tracking algorithms \cite{Kim21ICRA,Luiten2020,Qiao2020} consider object detection to be a prior step independent of tracking. In this paradigm, object detection is performed frame-by-frame. This runs contrary to how 3D data is collected using sensors (e.g. RGB camera, LiDAR) in settings like autonomous driving. Information about the detections in past frames can be an indicator of possible objects in future frames. However, almost all previous work in 3D object detection \cite{Lang_2019_CVPR,Shi_2020_CVPR,9018080,Yan2018} has focused on frame-by-frame detection. In our literature search, we find only Huang \etal\cite{Huang2020a} have tackled this issue previously. Ngiam \etal\cite{Ngiam2019} and Hu \etal\cite{Hu_2020_CVPR} also consider multiple 3D frames as input, but both use relatively simple techniques of reusing seed points or concatenating input over multiple frames.
\begin{figure}[t]
   \begin{center}
   \includegraphics[width=\linewidth]{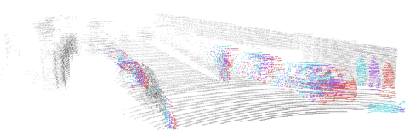}
   \end{center}
      \caption{Example sequence of frames in the KITTI\cite{Geiger2013IJRR} Tracking dataset. Objects in the same frame take the same color. A few cars, a  pedestrian and a cyclist are visible.}
   \label{fig:f2f_3d_objs}
\end{figure}

In this paper, we investigate a new method that makes use of temporal information across multiple LiDAR point clouds to simultaneously perform 3D detection and tracking. An example of the high correlation between object positions and object appearance in shown in Figure \ref{fig:f2f_3d_objs}, which visualizes 3 consecutive point clouds where objects in the same frame are colored similarly. Our objective is to continuously infer a 'tracklet' over multiple frames by correlating the object features for all detected Regions of Interest (RoIs). To this effect we propose to extend the PV-RCNN architecture proposed by Shi \etal\cite{Shi_2020_CVPR} with a tracking formulation which utilizes cross correlation to regress tracking values. We train an end-to-end architecture on a combined detection and tracking loss. We refer to this novel architecture as 3D-FCT for {\em3D Feature Correlation and Tracking}. The input to 3D-FCT consists of pairs of LiDAR point clouds which are first passed through a detection trunk (\eg PV-RCNN) to produce object features. These features are shared for both the detection and the tracking tasks. A cross-correlation is calculated between the features from the two adjacent point clouds. This correlation information is then used to predict the displacement of each object by regressing the box transformations across point clouds. Finally, we infer long-term tubes for objects using the short-term tracklets predicted by the network.

Compared to previous methods which concatenate input point clouds, our approach is more memory and compute efficient since the detection trunk is shared between the pairs of inputs as a Siamese network. Huang \etal\cite{Huang2020a} propose an LSTM-based architecture which also reduces memory and compute footprint. However, their architecture depends on the ability of the LSTM network to extract useful features from the {\em entire} point cloud as its hidden state for the next point cloud. 3D-FCT takes inspiration from the D2T architecture \cite{Feichtenhofer2017} in the 2D domain to utilize past features from each RoI separately.  An evaluation on the KITTI \cite{Geiger2013IJRR} Tracking Dataset shows that our approach is able to achieve better performance across all three classes as compared to frame-by-frame approaches, with an increase of 5.57\% mAP.

\section{Related Work}\label{sec:related_work}
\noindent{\bf 3D Object Detection.} Previous work on 3D object detection can be broadly divided into three categories: grid based methods, point based methods and representation learning. Grid based methods, pioneered by MV3D \cite{Chen2017}, dissect the 3D point cloud into a regular grid to apply a 2D or 3D CNN. The introduction of 3D sparse convolution \cite{Graham2018,Graham2017} allowed for efficient processing of voxels, leading to further development of 3D CNN architectures in \cite{9018080,Wang2019,Yan2018}. Point based methods use PointNet \cite{Charles2017,Qi2017} and its set abstraction operation to directly convert point clouds into feature vectors. F-PointNet \cite{Qi2018} and F-ConvNet \cite{Wang2019a} utilize 2D image bounding boxes to generate 3D proposals to be passed to PointNet. PointRCNN \cite{Shi2019} provides 3D proposals directly from point clouds. Finally, representation learning involves learning a set of features from the 3D point cloud before performing the object detection task. PointNet, proposed by Qi \etal \cite{Qi2017}, remains a key feature in representation learning, allowing flexible receptive fields of different search radii. The PV-RCNN \cite{Shi_2020_CVPR} model utilizes both voxel based feature learning and PointNet based feature learning to provide high quality detections.
\\\\
\noindent{\bf Spatio-temporal Methods.} A number of papers in the past few years have focused on using temporal information to improve 2D object detection in videos. This has been possible in large part because of video datasets like VidOR \cite{Shang2019,Thomee2016} which label object categories and relations over an extended sequence of image frames. Several methods utilize this temporal information only in the post-processing stage of the pipeline \cite{Han2016,Kang2017}, by constructing tubelets that are re-scored based on the object classes and 2D location. Feichtenhofer \etal \cite{Feichtenhofer2017} utilize a Siamese R-FCN detector to simultaneously detect and track objects in a video. FGFA \cite{Zhu2017} considers temporal information at the feature level, by utilizing optical flow from past and future frames to provide local aggregation. The MEGA \cite{Chen2020} network provides both local and global aggregation by allowing a large amount of information to be saved from past and future frames.

By contrast, using temporal information in LiDAR point clouds is largely an unexplored area of research. A direct approach followed by \cite{McCrae2020,Sallab2018,Yin2020} involves converting the point clouds into bird's-eye view projections and then utilize 2D CNN architectures along with ConvGRU or ConvLSTM. Luo \etal \cite{Luo2018} build a combined architecture for the related tasks of detection, tracking, motion forecasting and motion planning by concatenating multiple point clouds into a single input. Another related approach taken by Choy \etal \cite{Choy2019} builds a sparse 4D convolution with non-conventional kernels, with the 4th dimension being time. The approaches above can be considered as {\em early fusion} with the time dimension being added as an input. A {\em middle fusion} approach is taken with PointRNN \cite{Fan2019} which aggregates the past state and the current input based on point coordinates. Similarly, Huang \etal \cite{Huang2020a} use LSTM networks to preserve a hidden state from the previous inputs. Our approach is different as we do not use a neural network to pull features from the {\em entire} point cloud into a state vector. Instead, we correlate features of individual RoI across point clouds in time.
\\\\
\noindent{\bf 3D Object Tracking.} 
Previous effort has also focused on the related task of object tracking in 3D \cite{Kim21ICRA,Luiten2020,Qiao2020}. However, these works assume 3D object detection as a pre-existing input and do not necessarily improve it further. Another approach taken in FlowNet3D \cite{Liu2019} is to estimate per-point translation vectors to provide a 3D scene flow. Scene flow can provide information about the direction and magnitude of movement of objects, and hence might be useful for improving object detection.

\section{Method}\label{sec:method}
In this section we first provide a high level overview of 3D-FCT (Sect. \ref{sec:method:overview}) which outputs tracklets given two consecutive point clouds as input. In Sect. \ref{sec:method:DnT} we formulate the 3D object detection and tracking task using the PV-RCNN detector \cite{Shi_2020_CVPR} as a baseline. The tracking objective is formalized as a cross-cloud bounding box regression task (Sect. \ref{sec:method:multi_loss}) which is achieved using a correlation technique detailed in Sect. \ref{sec:method:keypoint_corr}.

The inter-cloud tracklets are combined into long-term tubes using the procedure described in Sect. \ref{sec:method:tracklets_to_tubes}. We finally apply 3D-FCT on the KITTI tracking dataset \cite{Geiger2013IJRR} in Sect. \ref{sec:results}.

\subsection{Overview}\label{sec:method:overview}
\begin{figure*}
   \begin{center}
   \includegraphics[width=\linewidth]{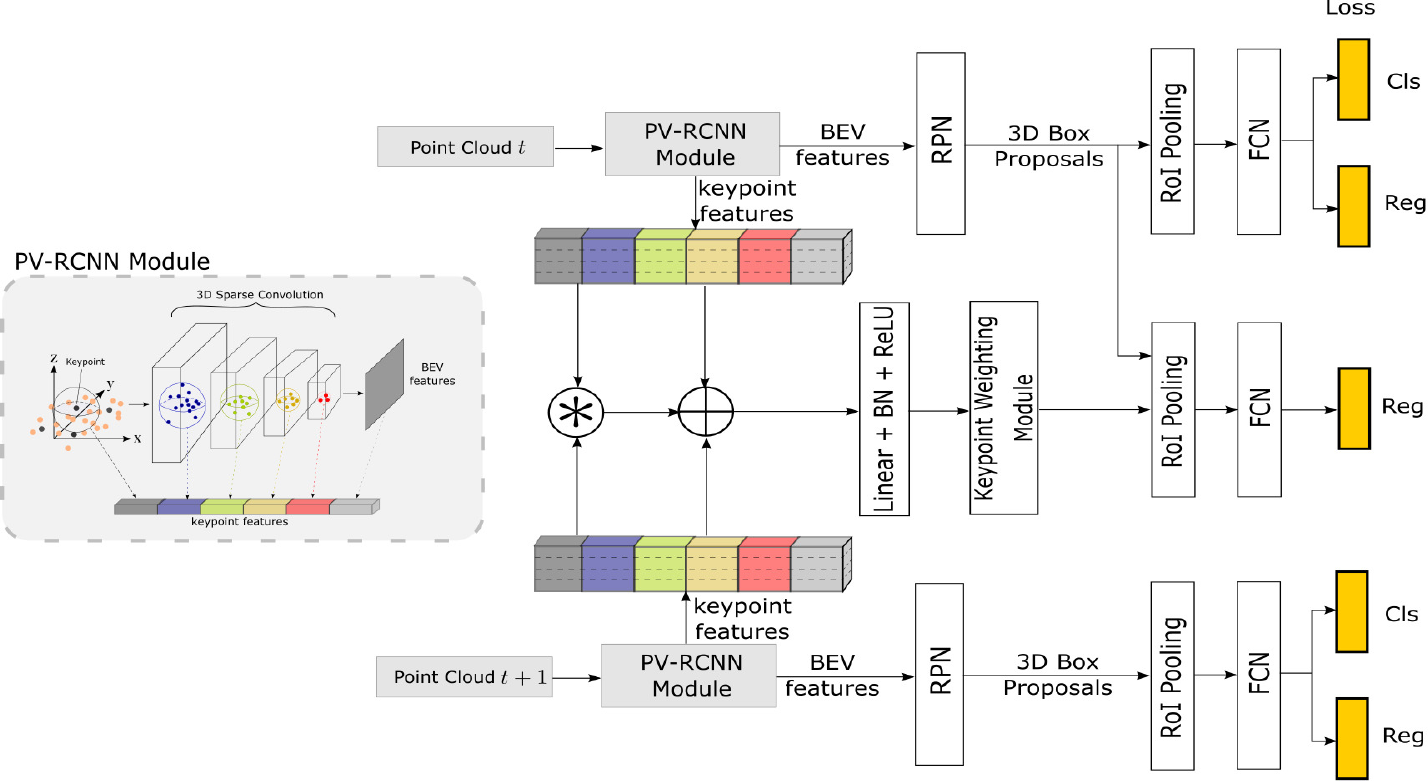}
   \end{center}
      \caption{Overview of the 3D-FCT temporal detection architecture. A pair of point clouds are processed by a Siamese PV-RCNN network to provide keypoint features. These keypoint features are compared using a correlation operation before being used to regress cross-cloud bounding box targets.}
   \label{fig:architecture}
\end{figure*}

We illustrate the overall architecture of the proposed 3D-FCT network in Fig. \ref{fig:architecture}. We use the PV-RCNN \cite{Shi_2020_CVPR} 3D object detection network which combines voxel based sparse 3D convolutions as well as the keypoint based representation learning to predict oriented bounding boxes and object classes. We extend this network for multi cloud detection and tracking. The {\em PV-RCNN module} as shown in Fig. \ref{fig:architecture} takes as input a point cloud and passes it through layers of 3D sparse convolution. In a parallel track, PV-RCNN selects {\em keypoints} using furthest point sampling and uses PointNet \cite{Qi2017} to convert the point cloud within a radius of the keypoint into a feature vector. In this manner, PV-RCNN is able to provide keypoint features across multiple layers (shown as different colors in Fig. \ref{fig:architecture}). Using these keypoint and BEV features, a RPN is used to propose candidate regions in each cloud based on pre-defined anchor boxes. An RoI pooling layer finally aggregates position-sensitive features and scores to classify objects and refine the box targets (regression).

The above architecture is extended by constructing a Siamese network \cite{Bromley1994} which accepts a set of point clouds as an input and each point cloud is passed through a PV-RCNN with shared weights. Further, a regressor is introduced which takes as input the keypoint features from both point clouds as well as a correlation map to predict the box transformations from one point cloud to the next. To achieve coherence across point clouds, features are pooled at the same proposal region as the point cloud at time $t$. The trunk of the network responsible for the tracking task is trained by introducing a new tracking regression loss into the multi-task objective of PV-RCNN. The tracking loss is defined as the smooth L1 norm between the predicted and ground-truth cross-cloud bounding box movement.

The formulation of tracking as the bounding box movement is common in 2D \cite{Held2016}, however, it does not inherently utilize translational equivariance. A correlation approach counters that shortcoming since correlation is robust to translations. A correlation map searches for feature in the search space that matches the template feature. In a single object tracking scenario, correlation maps can be utilized to calculate the displacement of an object by taking a maximum over the space.

Keypoint features extracted from PV-RCNN concatenate information from varying depths of the 3D convolution network allowing them to extract both low and high level features. Our aim is to develop a multi-object correlation tracker by computing correlation maps of the features of all keypoints in a point cloud. Our architecture can be trained end-to-end by using point clouds as an input to produce object detections and tracks. The next section describes how this end-to-end learning is formulated.

\subsection{3D object detection and tracking}\label{sec:method:DnT}
The 3D-FCT architecture takes as input a point cloud $\mathbf{P}^t\in \mathbb{R}^{m\times d}$ at time $t$. This point cloud is first divided into small voxels of size $L\times W\times H$ with non-empty voxels taking the mean value of all point-wise features for the points that fall in the voxel. As in PV-RCNN \cite{Shi_2020_CVPR}, we utilize a series of $3\times 3\times 3$ 3D sparse convolution layers to downsample the input by $1\times, 2\times, 4\times$ and $8\times$ respectively. 3D bounding box proposals are then calculated using an anchor-based approach from a BEV feature map of dimension $\frac{L}{8}\times\frac{W}{8}$ after stacking the 3D features along the $Z$ axis. As a result we have a bank of $D_{cls} = 2\times (C+1)\times\frac{L}{8}\times\frac{W}{8}$ position-sensitive score maps for two anchors at $0\degree, 90\degree$ for $C$ classes (and background) at every pixel of the BEV map. Using a second regression branch we also have a bank of $D_{reg} = 7\times\frac{L}{8}\times\frac{W}{8}$ class-agnostic bounding box predictions of a 3D box $b = (x, y, z, l, w, h, r_z)$.

In a parallel trunk, PV-RCNN \cite{Shi_2020_CVPR} selects $n$ keypoints $\mathcal{K} = \{p_1, \ldots, p_n\}$ from the point cloud  $\mathbf{P}^t$, where $n=2,048$ for the KITTI dataset. A voxel set abstraction module subsequently extracts keypoint specific features from each level of the 3D voxel CNN. These features are concatenated together to provide a multi-scale semantic feature $f_i = \left[f_i^{(pv_1)}, f_i^{(pv_2)}, f_i^{(pv_3)}, f_i^{(pv_4)}\right]$ for each keypoint $p_i$.

We now consider a pair of point clouds $\mathbf{P}^t, \mathbf{P}^{t+\tau}$ collected at time $t$ and $t+\tau$ as inputs to the network. An inter-cloud bounding box regression layer is introduced that performs position sensitive RoI pooling on a concatenation of the multi-scale semantic features $\{f^t, f^{t+\tau}\}$ as well as the correlation between them (described in more detail in Sect. \ref{sec:method:keypoint_corr}). The regression layer is trained to predict the transformation $\Delta^{t+\tau} = (\Delta_{x}^{t+\tau}, \Delta_{y}^{t+\tau}, \Delta_{z}^{t+\tau}, \Delta_{r_z}^{t+\tau})$ of the RoIs from $t$ to $t+\tau$.

\subsection{Multitask detection and tracking objective}\label{sec:method:multi_loss}
PV-RCNN \cite{Shi_2020_CVPR} is trained end-to-end on a combination of classification loss $L_{cls}$ and regression loss $L_{reg}$. A tracking loss $L_{tra}$ is added to this overall loss function to allow the tracking regressor to be trained. Given $N$ RoIs with softmax probabilities $\{p_i\}_{i=1}^N$, regression offsets $\{b_i\}_{i=1}^N$ and cross cloud RoI tracking transformations $\{\Delta_i^{t+\tau}\}_{i=1}^{N_{tra}}$, the overall loss function for a single iteration can be written as:
\begin{align}
   \label{eq:loss}
   L(\{p_i\}, \{b_i\}, \{\Delta_i^{t+\tau}\}) = \frac{1}{N}\sum_{i=1}^{N}L_{cls}(p_{i,c^*})\nonumber\\
   +\lambda_{reg}\frac{1}{N_{fg}}\sum_{i=1}^{N} [c_i^*>0]L_{reg}(b_i, b_i^*)\\
   + \lambda_{tra}\frac{1}{N_{tra}}\sum_{i=1}^{N_{tra}}L_{tra}(\Delta_i^{t+\tau}, \Delta_i^{*, t+\tau})\nonumber
\end{align}

The classification loss is calculated as the cross entropy loss, $L_{cls}(p_{i,c^*}) = -\log{(p_{i,c^*})}$ where $p_{i,c*}$ is the predicted softmax score for the ground-truth label $c^*$. Hence, the first term in eq. \ref{eq:loss} acts for all $N$ RoIs. The regression loss (second term in eq. \ref{eq:loss}) acts only for $N_{fg}$ foreground RoIs. The indicator function $[c_i^*>0]$ is $1$ for RoIs with $c_i^*\neq 0$ and $0$ for RoIs with $c_i^*=0$. $L_{reg}$ is calculated as the smooth L1 loss defined in \cite{Girshick2015} between the predicted bounding boxes $b_i$ and the ground truth bounding boxes $b_i^*$. Assignment of RoIs to ground truth is based on a minimum intersection-over-union (IoU) of $0.5$. The third term in eq. \ref{eq:loss} corresponds to the tracking loss, and is active only for foreground RoIs for which the ground truth object is present in both point clouds $\mathbf{P}^t$ and $\mathbf{P}^{t+\tau}$. Therefore, we have $N_{tra}\leq N_{fg}\leq N$. The tradeoff parameters $\lambda_{reg}$ and $\lambda_{tra}$ were both set to $1$ for this work.

The ground truth labels for track regression are calculated based on the ground truth bounding boxes in both point clouds. The bounding box for a single object in point cloud $\mathbf{P}^t$ can be defined as $b^t = (x^t, y^t, z^t, l, w, h, r_z^t)$ denoting the 3D coordinates of the center, length, width, height and yaw angle respectively. Consider the corresponding bounding box $b^{t+\tau}$ for the same object in time $t+\tau$. The target for the tracking regression $\Delta^{*,t+\tau} = (\Delta_{x}^{*,t+\tau}, \Delta_{y}^{*,t+\tau}, \Delta_{z}^{*,t+\tau}, \Delta_{r_z}^{*,t+\tau})$ is then defined as:
\begin{align}
   \label{eq:tracking_gt}
   \Delta_{x}^{*,t+\tau} = \frac{x^{t+\tau} - x^t}{l} \quad \Delta_{y}^{*,t+\tau} = \frac{y^{t+\tau} - y^t}{w}\\
   \Delta_{z}^{*,t+\tau} = \frac{z^{t+\tau} - z^t}{h} \quad \Delta_{r_z}^{*,t+\tau} = \frac{r_z^{t+\tau} - r_z^t}{2\pi}
\end{align}
Note that the length, width and height of an object does not change over time in 3D object detection.

\subsection{Keypoint correlation for object tracking}\label{sec:method:keypoint_corr}
\begin{figure*}
   \begin{center}
   \includegraphics[width=\linewidth]{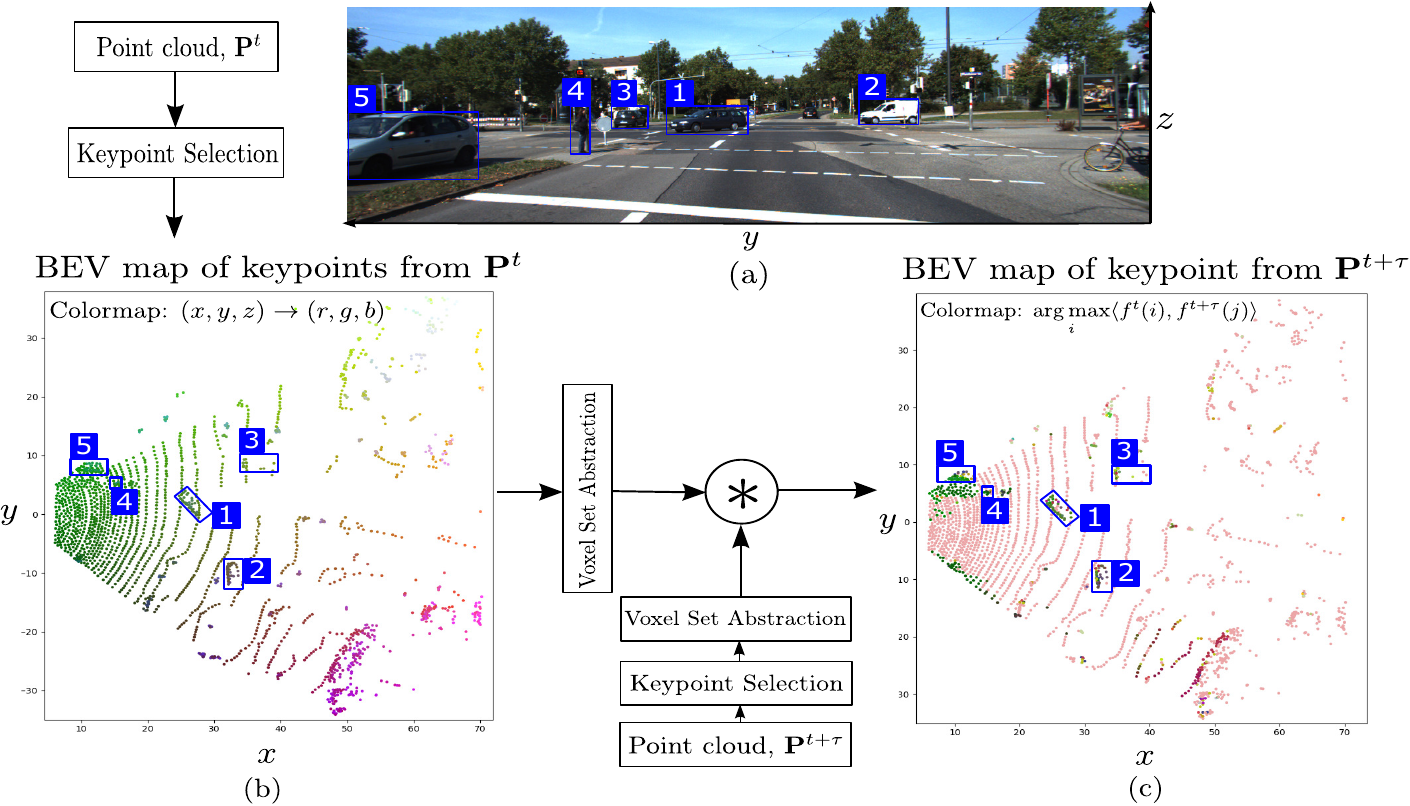}
   \end{center}
      \caption{Correlation features for two point clouds from the validation dataset (Seq 2, clouds 134 \& 135). (a) RGB image at time $t$ for comparison. (b) BEV view of keypoints from point cloud $\mathbf{P}^t$, with RGB color mapped from the $(x,y,z)$ location. (c) BEV view of keypoints from point cloud $\mathbf{P}^{t+\tau}$ with color mapped to the keypoint color in (b) that maximizes the correlation. We see that the correlation between keypoints of objects 1 to 5 (correspondingly labelled in (a)) is fairly accurate. Moreover, the correlation for keypoints belonging to the road surface is identically maximal for the same point in $\mathbf{P}^t$. This can be attributed to the fact that a local feature descriptor for road surface is similar regardless of global location. This also highlights the potential use of the proposed architecture in segmentation applications. Best viewed in color.}
   \label{fig:correlation}
\end{figure*}

Defining correlation in 3D space is challenging because of the sparsity of 3D data. The simplest method would be to define correlation over the voxelized 3D space. This technique is already quite memory intensive in 2D space, in 3D space it would be prohibitive. Furthermore, unlike in the case of single object tracking, multi object tracking requires construction of correlation maps over multiple templates. Considering all possible circular shifts in 3D voxel space would produce feature maps of very large sizes. Hence, we utilize the multi-scale semantic keypoint features $\{f^t, f^{t+\tau}\}$ provided by PV-RCNN. The correlation layer in Fig. \ref{fig:architecture} performs keypoint-wise feature comparison between keypoints in two point clouds:
\begin{align}\label{eq:feature_corr}
   f_{corr}^{t,t+\tau}(i,j) = \langle f^t(i), f^{t+\tau}(j)\rangle\quad \forall i,j\in \{1,\ldots,n\} 
\end{align}
Thus the correlation layer produces a feature map $f_{corr}^{t,t+\tau}\in\mathbb{R}^{n\times n}$. Since the correlation feature is calculated from the keypoint features, it takes into account features from various semantic scales of the 3D sparse convolutional network.

The correlation features are concatenated with the keypoint features as $\{f_{corr}^{t,t+\tau}, f^t, f^{t+\tau}\}$ before being passed through a fully connected layer to produce the final features $f_{comb}\in\mathbb{R}^{n\times n_d}$, with $n_d=256$ for this work. These features are weighted based on the keypoint weighting module of PV-RCNN before they are pooled by the RoI pooling layer. Fig. \ref{fig:correlation} shows an example of the correlation map generated by the correlation of keypoint features. In Fig \ref{fig:correlation}(b), the keypoint of the point cloud at time $t$ are visualized in a bird eye view (BEV) image, with the color of the points mapped to their respective $(x,y,z)$ location. Fig \ref{fig:correlation}(c) correspondingly visualizes the keypoints of cloud at time $t+\tau$. Here however the points are colored to match the keypoint in cloud $t$ that maximizes the correlation. We observe that for most objects (highlighted in both Fig. \ref{fig:correlation}(a)\&(c)) the correlation is fairly accurate, maintaining the correspondence across time. We also observe that the road plane is mapped to a single point that maximizes the correlation. This is because localized features of a road surface are expected to be very close regardless of position, and the correlation function does not consider global positions. However, it is intersting to see that patches of grass are accurately correlated across the two point clouds, which highlights the possible future use case scenario of this work for point cloud segmentation.

\subsection{Linking tracklets to object tubes}\label{sec:method:tracklets_to_tubes}
The high dimensionality of point clouds put a constraint on the number of point clouds that can be simultaneously processed by a single network based on the limitations of the memory available in modern GPUs. This is exacerbated by the fact that 3D object detection is commonly applied in autonomous robotics applications where processing ability may be limited. Hence, a method is required to combine the cross cloud tracklets that are predicted by our method into long-term tracks for individual objects. To this end, we utilize a location based technique, similar to those used in \cite{Feichtenhofer2017,Han2016} in 2D, to link detections into long term tracks.

Consider the detections for a point cloud at time $t$, $B_{i,c}^{t} = \{x_i^t, y_i^t, z_i^t, l_i, w_i, h_i, r_{z,i}^t, p_{i,c}^t\}$, where $B_{i,c}^{t}$ is a bounding box indexed by $i$, centered at $(x_i^t, y_i^t, z_i^t)$ with dimensions $(l_i, w_i, h_i)$ and a softmax probability of $p_{i,c}^t$ for class $c$. In addition, we have predictions for the position of object $B_{i,c}^t$ at time $t+\tau$, i.e. the tracklets $T_i^{t, t+\tau} = \{x_i^t+\Delta_{x}^{t,t+\tau}, y_i^t+\Delta_{y}^{t,t+\tau}, z_i^t+\Delta_{z}^{t,t+\tau}, l_i, w_i, h_i, r_{z,i}^t+\Delta_{r_z}^{t,t+\tau}\}$. Using these predictions we can define a class specific linking score allowing us to combine detections across time
\begin{align}\label{eq:link_score}
   s_c(B_{i,c}^t, B_{j,c}^{t+\tau}, T^{t,t+\tau}) = p_{i,c}^t + p_{j,c}^{t+\tau} + \phi(B_j^{t+\tau}, T^{t,t+\tau})
\end{align}
where the pairwise score is
\begin{align}\label{eq:pairwise_score}
   \phi(B_j^{t+\tau}, T^{t,t+\tau})) = \begin{cases}
      1,\quad\text{if } IoU(B_j^{t+\tau}, T^{t,t+\tau})>\delta)\\
      0,\quad\text{otherwise}
   \end{cases}
\end{align}
where $\phi$ evaluates to 1 if 3D volumetric IoU between the expected bounding box and the predicted bounding box at time $t+\tau$ is greater than the threshold $\delta$.
\begin{figure}[t]
   \begin{center}
   \includegraphics[width=0.8\linewidth]{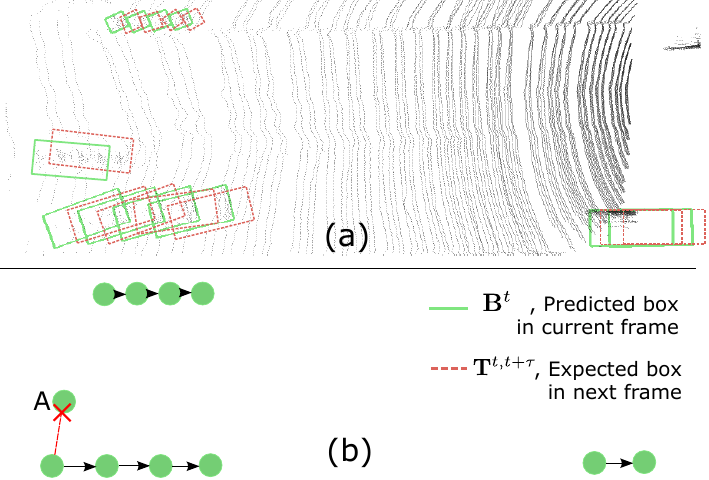}
   \end{center}
      \caption{(a) Four consecutive point cloud BEVs superimposed along with the predicted boxes $\mathbf{B}^t$ and the expected boxes in the next frame $\mathbf{T}^{t,t+\tau}$. (b) Illustration of linking tracklets into long term tracks based on $IoU(\mathbf{B}^{t+\tau}, \mathbf{T}^{t,t+\tau})$. Object A is removed from the predicted objects as it does not have $IoU>0$ with any past/future point clouds.}
   \label{fig:seqnms}
\end{figure}

A set of graphs are constructed where bounding boxes $\mathbf{B}^{t_0}$ is connected to $\mathbf{B}^{t_0+\tau}$ with edges weighted by $s_c(\mathbf{B}^{t_0}, \mathbf{B}^{t_0+\tau}, \mathbf{T}^{t_0,t_0+\tau})$, $\forall t_0\in(1, \ldots,T-\tau)$. Given such a set of graphs, our aim is to select the sequence of boxes that maximize the sum of the weights along the sequence. This can be done using a dynamic programming algorithm that maintains the maximum scoring sequence seen so far at each box. Once such a maximizing sequence $\mathbf{B}_{seq} = \{B_{i_s}^{t_s}, \ldots, B_{i_e}^{t_e}\}$ and its respective score sequence $\mathbf{S}_{seq'} = \{s_{i_s}^{t_s}, \ldots, s_{i_e}^{t_e}\}$ have been found, the scores within the sequence are updated using a function $\mathbf{S}_{seq} = F(\mathbf{S}_{seq'})$. For this work, we consider $F$ to be the maximum function. The aim is to boost the scores for positive boxes on which the detector fails to achieve sufficiently high scores. The boxes in $\mathbf{B}_{seq}$ are removed from the graph along with any boxes in $\mathbf{B}_t, t\in [t_s, t_e]$ with an IoU with $\mathbf{B}_{seq}^t$ greater than a set threshold. This process is repeated until no more links can be formed. This is summarized in Fig. \ref{fig:seqnms} which shows the BEV of 4 consecutive point clouds in time superimposed on each other. In Fig. \ref{fig:seqnms}(b), 3 long term tracks are formed based on eq. \ref{eq:link_score}. Such an approach prevents us from assuming that the object stays in view for the entirety of the point cloud sequence, which is seldom the case in 3D.

\section{Experiments}\label{sec:results}
\begin{table*}
   \begin{center}
   \resizebox{\textwidth}{!}{
   \begin{tabular}{l|ccccccccc|ccc}
   \hline
                                     & \multicolumn{3}{c}{Car}                         & \multicolumn{3}{c}{Pedestrian}                  & \multicolumn{3}{c|}{Cyclist}                     & \multicolumn{3}{c}{mAP}  \\
   Methods                           & \textit{Easy} & \textit{Medium} & \textit{Hard} & \textit{Easy} & \textit{Medium} & \textit{Hard} & \textit{Easy} & \textit{Medium} & \textit{Hard} & Easy & Medium & Hard     \\ 
   \hline
   PointPillar \cite{Lang_2019_CVPR} & 75.28 & 51.54 & 49.64 & 22.08 & 23.12 & 23.07 & 68.27 & 55.55 & 54.81 & 55.21 & 43.40 & 42.51\\
   SECOND \cite{Yan2018} & 85.33 & 69.04 & 67.17 & 46.73 & 47.08 & 46.87 & 78.89 & 68.96 & 68.26 & 70.32 & 61.69 & 60.77\\
   PointRCNN \cite{Shi2019} & 83.75 & 65.03 & 65.08 & 45.14 & 44.43 & 42.44 & 55.30 & 43.22 & 41.27 & 61.40 & 50.89 & 49.60\\
   \hline
   PV-RCNN \cite{Shi_2020_CVPR} & 84.54 & 72.07 & 70.57 & 48.99 & 52.35 & 52.60 & 82.63 & 70.83 & 70.36 & 72.05 & 65.08 & 62.75 \\
PV-RCNN + T loss       & 86.11 & \textbf{72.79} & 71.28 & 49.05 & 53.32 & 52.11 & 83.66 & 74.22 & 74.31 & 72.94 & 66.78 & 65.90 \\
3D-FCT (Ours) & \textbf{89.56} & 72.07 & \textbf{72.72} & \textbf{55.52} & \textbf{58.40}  & \textbf{58.94} & \textbf{89.15} & \textbf{75.86}  & \textbf{77.49} & \textbf{78.08} & \textbf{68.78} & \textbf{69.72}  \\
\rowcolor{LightCyan}
\textit{Improvement} & \textit{+5.02} & \textit{+0.00}  & \textit{+2.15} & \textit{+6.53} & \textit{+6.05}  & \textit{+6.34} & \textit{+6.52} & \textit{+5.03}  & \textit{+7.13} & \textit{+6.03} & \textit{+3.70}  & \textit{+6.97} 
\\
\hline   
\end{tabular}}
\end{center}
\caption{Performance comparison on KITTI tracking validation set. The average precision (in \%) for each class and each difficulty category is shown, along with the mean average precision over all classes. We use PV-RCNN \cite{Shi_2020_CVPR} as a backbone for this architecture.\label{table:evaluation}}
\end{table*}

\subsection{Dataset and Evaluation}
Our 3D-FCT architecture is evaluated on the KITTI \cite{Geiger2013IJRR} tracking dataset which is used for multi object tracking evaluation for autonomous vehicle applications. The dataset consists of three classes: cars, pedestrians and cyclists, with 21 training sequences and 29 test sequences. Each sequence consists of consecutive point clouds collected over a period of time. Since the test dataset does not provide ground truth information, the training sequences are divided into a training (5027 point clouds) and validation (2981 point clouds) set, respectively. In this, we follow \cite{Voigtlaender2019} who split the dataset to approximately balance the relative number of occurrence of each class equally across the two sets\footnote{Sequences 0, 1, 3, 4, 5, 9, 11, 12, 15, 17, 19, 20 were used for training, while the remaining were chosen for validation.}. Similar to the standard practice, as presented in \cite{Simonelli2020} and used by the KITTI 3D object detection challenge, we utilize the 3D average precision (AP) using 40 recall positions to evaluate our results. In addition, a true positive requires a minimum 3D overlap of 70\% for cars and 50\% for pedestrians and cyclists.

For all evaluations, we begin with a model pretrained on the KITTI \cite{Geiger2013IJRR} 3D object detection dataset before training it on the tracking dataset. Also, we do not follow the data augmentation technique of oversampling ground truth objects as described in \cite{Cheng2020,Yan2018} and common practice with most 3D object detection neural network implementations. This augmentation technique places ground truth objects at random locations in the point cloud, greatly increasing the number of objects available for training and also solving the class imbalance. However, our implementation requires the existence of short term tracklets for each object between consecutive frames which makes the use of this augmentation technique challenging. Extension of this technique for tracking formulations is left for future work.

\subsection{Training}
\noindent{\bf PV-RCNN.} Our PV-RCNN backbone is trained exactly as originally proposed in \cite{Shi_2020_CVPR}. The 3D voxel CNN consists of 4 level with feature dimensions 16, 32, 64, 64 respectively. The radii levels used in the voxel set abstraction module are also preserved from \cite{Shi_2020_CVPR} as (0.4m, 0.8m), (0.8m, 1.2m), (1.2m, 2.4m), (2.4m, 4.8m) for each layer respectively, with the neighborhood radii for the set abstraction of for raw points being (0.4m, 0.8m). The voxel size was set to be (0.1m, 0.1m, 0.15m). The RoI grid pooling operation samples $6 \times 6 \times 6$ grid points in each 3D proposal. The number of keypoints is fixed to $n=2048$. The generated proposals are suppressed to 512 for training and 100 for inference using NMS.

\noindent{\bf Tracking.} For training the tracking arm of 3D-FCT, we start with a pretrained PV-RCNN model and fine tune it based on the added tracking loss from eq. \ref{eq:loss}. For all of our evaluations we utilize a value of $\tau = 1$, which means we use directly consecutive point clouds as input. This is primarily because LiDAR point clouds are collected at a much lower frequency ($\sim$10 clouds per second) than RGB images (30/60 FPS). As a result, objects can move by significant amounts from 1 point cloud to the next. The feature convolution operation in eq. \ref{eq:feature_corr} is performed over all $n=2048$ keypoints. We recognize that this may not be necessary and a more efficient strategy would be to only perform correlation with keypoints within a distance of the current keypoint. However, we leave an analysis on the benefits of this approach to future work. The multi-scale semantic features retrieved from the voxel set abstraction module is of length $128$. Hence, after concatenating the correlation features $f_{corr}$, the sematic features $f$ and the global locations of the keypoints, the overall size of the features for a single iteration is $2048\times 2310$. This size is reduced to $2048 \times 256$ using a single fully connected layer. After the RoI pooling operation, the features are passed through a fully connected network of 4 layers of 256 neurons each for the regression task.

\subsection{Results}
\begin{table}
   \centering
   \resizebox{0.49\textwidth}{!}{
   \begin{tabular}{cc|ccc|c}
   \hline
      Track      & Tracklet & \multicolumn{4}{c}{AP (\textit{Moderate difficulty}, \%)}                                                              \\
      Regression & Linking  & \multicolumn{1}{l}{Car} & \multicolumn{1}{l}{Pedestrian} & \multicolumn{1}{l|}{Cyclist} & \multicolumn{1}{l}{mAP}  \\ 
   \hline
   \xmark & \xmark & 72.07 & 52.35 & 70.83 & 65.08\\
   \xmark & \cmark & 43.23 & 53.85 & 72.37 & 56.48\\
   \cmark & \xmark & \textbf{72.79} & 53.32 & 74.22 & 66.78\\
   \cmark & \cmark & 72.07 & \textbf{58.40} & \textbf{75.86} & \textbf{68.78}\\
   \hline                     
   \end{tabular}}
   \caption{Ablation study on effect of track regression and tracklet linking. PV-RCNN \cite{Shi_2020_CVPR} acts as the baseline.}
   \label{tab:ablation}
\end{table}

We perform a series of experiments to evaluate the performance of our proposed network against alternative approaches. We also perform an ablation study to identify the effect of each sub-component of our network. These results are summarized in Table \ref{table:evaluation}.

Our first baseline is the PV-RCNN \cite{Shi_2020_CVPR} model that was trained on the KITTI tracking dataset, which achieves 65.08\% mAP for the medium difficult category. Given that we do not utilize the data augmentation technique mentioned in \cite{Yan2018}, the AP for the 'Car' category is much higher than that for the 'Pedestrian' category, reflecting the imbalance inherent in the data. Our second baseline aims to measure the benefit of adding the tracking loss on the overall AP for 3D object detection. Adding this loss based on the track regression features provides a mAP of 66.78\% (an increase of 1.7\% over the baseline). We believe that the tracking loss improves the mAP since it reaffirms the important objects in the training data and acts as a regularizer pushing the features of the same object closer together.

Next, we evaluate the complete proposed architecture combining the tracking loss as well as the use of the regressed track predictions to link tracklets into long term tracks (3D-FCT). This leads to significant increase in performance with a mAP of 68.78\% for the medium difficulty objects, an increase of 3.7\% from the baseline. We see especially large improvements for the pedestrian and cyclist categories (+6.05\% and +5.03\% respectively). The proposed 3D-FCT architecture achieves this improvement by detecting objects that could have potentially been missed by the baseline due to occlusion, irregular pose or the fact that objects farther away from the sensor have less descriptive points. This is also why we see larger performance improvements for cyclists and pedestrians, which are smaller than cars, described by fewer points, and have higher variability in  poses.

We also report the performance of other single cloud detectors evaluated on the KITTI tracking dataset. The PointPillars \cite{Lang_2019_CVPR} model is able to achieve only 43.4\% mAP for the medium difficulty category.  We observe that PointPillars is not very robust to class imbalance in the dataset and achieves only 23.12\% AP for the pedestrian class. SECOND \cite{Yan2018} performs better than both PointPillars \cite{Lang_2019_CVPR} and PointRCNN \cite{Shi_2020_CVPR}, achieving 61.69\% mAP. The proposed model is able to outperform both the PV-RCNN baseline as well as other single cloud detectors on the KITTI tracking evaluation set.

In Table \ref{tab:ablation}, we perform an ablation study to identify the effect of each component of the proposed architecture. The table shows the average precision for each class for the moderate difficulty objects from the validation KITTI tracking set. Note that if linking tracklets into long term tracks is done without track regression (based on $IoU(\textbf{B}^t, \textbf{B}^{t+\tau})$ rather than $IoU(\textbf{T}^{t,t+\tau}, \textbf{B}^{t+\tau})$), a sharp drop in performance for the car category is observed. This drop in performance is largely due to the increase in false positives that are predicted by the neural network. Objects can move by significant proportions of their length in the course of a single time step. Hence, it is imperative to combine the track regression along with the tracklet linking to achieve best results.

\section{Discussion}\label{sec:discussion}
Our proposed 3D-FCT architecture aims to utilize temporal information to perform simultaneous 3D object detection and tracking. We leverage the correlation of keypoints in the feature space to predict the tracking regression of objects from time $t$ to time $t+\tau$. These predicted track regression values allow the prediction of long term tracks by linking inter-cloud tracklets across time. Experiments on the KITTI tracking dataset demonstrate that our proposed method is able to improve the mAP by 5.57\% (averaged across the difficulty categories). In the future, we would also like to extend the architecture to produce 3D object segmentation maps, as the keypoint correlation information can be used directly for that purpose. Moreover, we would like to study the effect of computing the correlation between keypoints within a certain distance from each other, as opposed to calculating them for all possible pairs.

\textbf{Memory Efficiency:} Although the 3D-FCT architecture in Fig. \ref{fig:architecture} depicts two point clouds as an input to the network, both halves of the siamese network are independent of each other. Therefore, a more efficient implementation would only accept point cloud $\mathbf{P}^{t+\tau}$ as an input, and utilize the keypoint features $f^t$ saved from the previous iteration to predict the track regressions. Such an implementation would be more memory efficient that other approaches that concatenate point clouds to process them together \cite{Lang_2019_CVPR,Luo2018}. We also note that since the correlation is performed on extracted keypoints only, the correlation workload is lightweight in comparison to performing the same over all possible points.

\textbf{Online capabilities:} Online application is only limited by the linking on object tracklets across time. The formulation defined in Sec \ref{sec:method:tracklets_to_tubes} combines tracklets based on both past and future point clouds. To allow for online track formation, this can be changed to consider only causal effects at time $t_0\in(t-\omega, t)$ where $\omega$ is the effective window of time over which objects are tracked.

\section{Conclusion}\label{sec:conclusion}
We have presented a novel 3D-FCT architecture to perform simultaneous 3D object detection and tracking on LiDAR points clouds. Our proposed method allows for end-to-end training in a memory efficient manner. Our approach has been shown to provide improvements in average precision on the KITTI tracking dataset over other baseline approaches. Jointly performing object detection and tracking has been shown to be effective in 2D in research, and we demonstrate that this approach is useful in 3D domain as well. We hope that this work can encourage other research in the same area.


{\small
\bibliographystyle{ieee_fullname}
\bibliography{paper}
}

\end{document}